\documentclass{article}

\usepackage{spconf,amsmath,graphicx}
\usepackage{xcolor}
\usepackage{booktabs}
\usepackage{bbding}
\usepackage{hyperref}
\usepackage{amssymb} 
\usepackage[numbers]{natbib}
\usepackage{stmaryrd}
\usepackage{caption}
\usepackage{subcaption}
\usepackage{graphicx}
\title{IMAGE CLASSIFICATION FOR CSSVD DETECTION IN CACAO PLANTS}

\name{Atuhurra Jesse, N'guessan Yves-Roland Douha, Pabitra Lenka }
\address{ \{atuhurra.jesse.ag2; douha.nguessan\_yves-roland.dn6; lenka.pabitra.lm3\} @is.naist.jp \\ Nara Institute of Science and Technology}

\begin{document}
\topmargin=0mm 
\maketitle
\begin{abstract}
The detection of diseases within plants has attracted a lot of attention from computer vision enthusiasts. Despite the progress made to detect diseases in many plants, there remains a research gap to train image classifiers to detect the cacao swollen shoot virus disease or CSSVD for short, pertinent to cacao plants. This gap has mainly been due to the unavailability of high quality labeled training data. Moreover, institutions have been hesitant to share their data related to CSSVD. To fill these gaps, we propose the development of image classifiers to detect CSSVD-infected cacao plants. Our proposed solution is based on VGG16, ResNet50 and Vision Transformer (ViT). We evaluate the classifiers on a recently released and publicly accessible KaraAgroAI Cocoa dataset. Our best image classifier, based on ResNet50, achieves 95.39\% precision, 93.75\% recall, 94.34\% F1-score and 94\% accuracy on only 20 epochs. There is a +9.75\% improvement in recall when compared to previous works. Our results indicate that the image classifiers learn to identify cacao plants infected with CSSVD.
\end{abstract}
\section{Introduction}
\label{sec:intro}
\begin{figure*}
     \centering
     \begin{subfigure}[t]{0.38\textwidth}
         \centering
         \includegraphics[width=\textwidth, height=5.0cm]{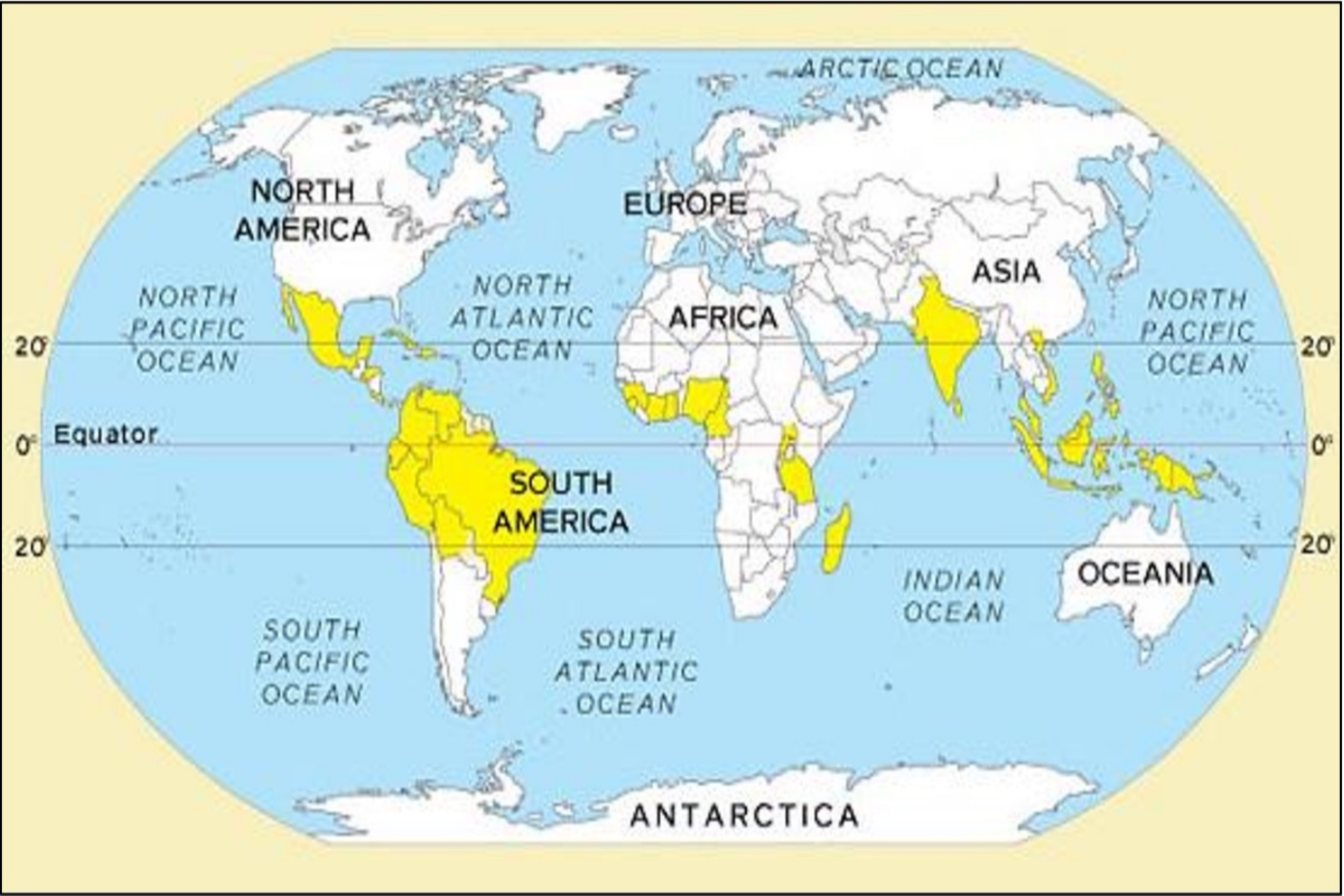}
         \caption{Countries where CSSVD is prevalent are shown on the map in yellow.}
         \label{fig:cssvd_map}
     \end{subfigure}
     \hfill
     \begin{subfigure}[t]{0.58\textwidth}
         \centering
         \includegraphics[width=\textwidth, height=5.0cm]{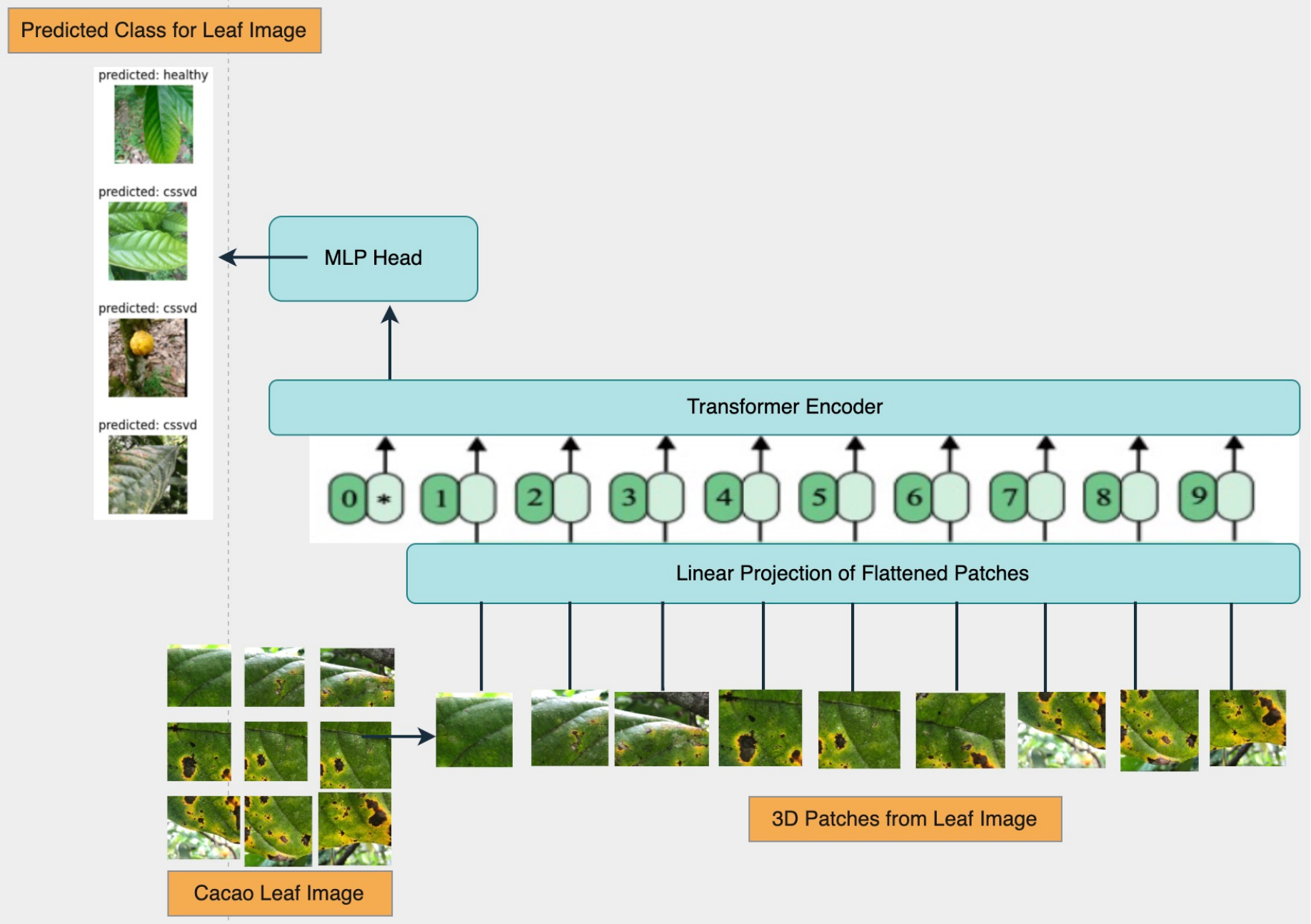}
         \caption{An image classifier based on the Vision Transformer (ViT) for CSSVD detection.}
         \label{fig:ViT}
     \end{subfigure}
    \caption{ \textit{Left:} CSSVD prevalence. \textit{Right:}  CSSVD detection with a Vision Transformer.}
    \label{fig:cssvd_map_and_ViT}
\end{figure*}

The ability to recognize patterns in images and solve real-world problems is one of the main goals of computer vision researchers. Contemporary image classification includes training a neural network to learn the patterns within an image. From such patterns, the neural network distinguishes images that belong to different classes. Image classification has been dominated by CNN-based methods. The detection and classification of diseases within plants has attracted a lot of attention from computer vision enthusiasts, not only because they can train already available image classifiers to detect infected plants, but they can also contribute to a greater good, say, by enriching the lives of farmers. This work tackles the CSSVD disease which affects many farmers around the world. Figure~\ref{fig:cssvd_map} depicts countries where CSSVD is a major problem. Next we review recent studies that leverage image classifiers to identify diseases in plants.
%%%%%%%%%%%%%%%%%%%%%%%%%%%%%%
\begin{figure}[t]
\centering
\includegraphics[width=8.5cm, height=4.5cm]{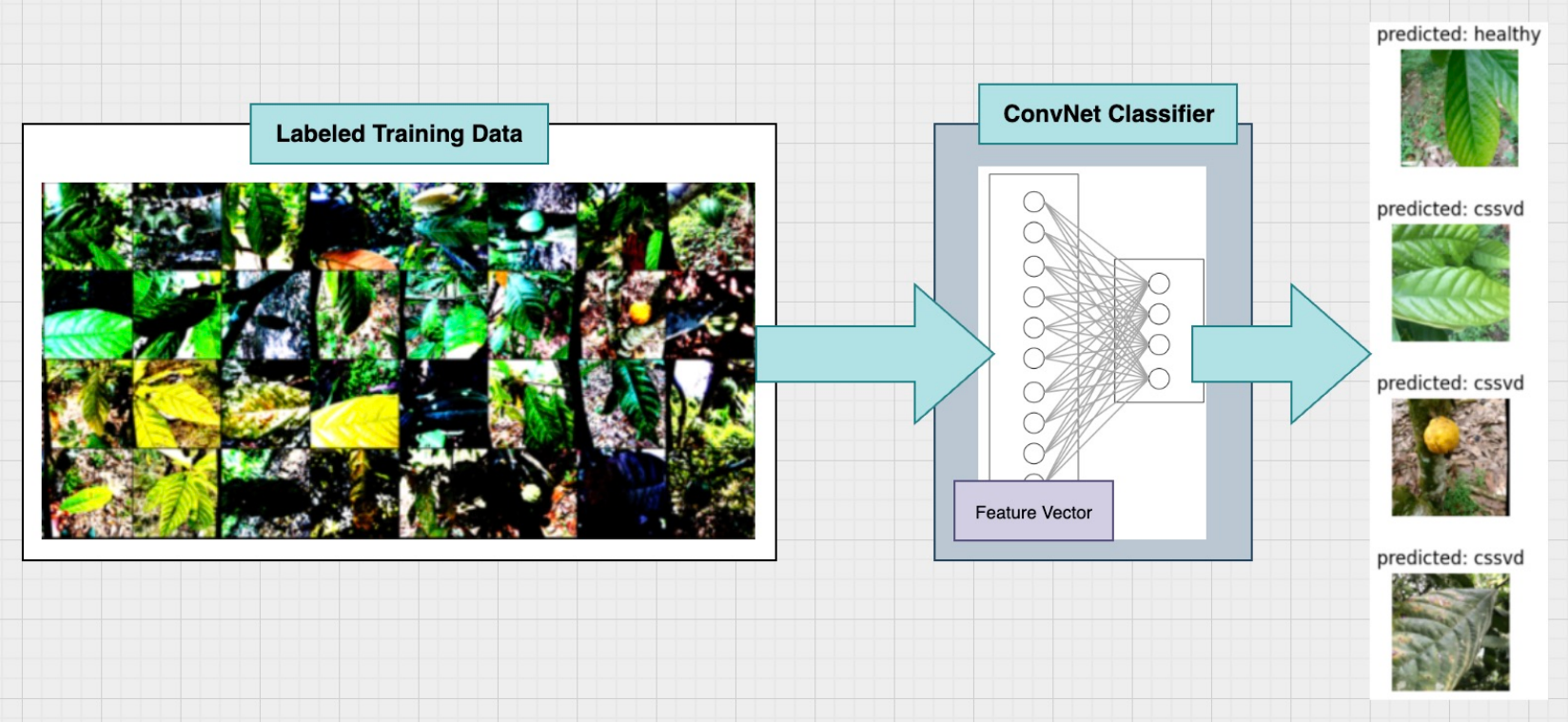}
\caption{ConvNet classifiers for CSSVD detection. Two ConvNets are used in this experiment. That is, VGG16 and ResNet50. In both cases, feature vectors are extracted from images of cacao leaves, stems and fruits. The classifier is then trained to extract distinguishing features to categorize an image into one of three classes, namely: Anthracnose, CSSVD, Healthy.}
\label{fig:ConvNet}
\end{figure}
%%%%%%%%%%%%%%%%%%%%%%%%%%%%%%

A recent study by~\citep{agriculture12081192} presents the methods, datasets and applications of image classification to plants. In this study, it is eminent that despite CSSVD being a major problem, it has not received a lot of researchers' attention. Specifically, this study presents no works at all about the detection of diseases in cacao plants. This research gap is in part due to lack of high quality labeled data necessary to train image classifiers to distinguish between infected and non-infected cacao plants.

To fill this gap, we propose the development of image classifiers to detect cacao plants infected with CSSVD. Our proposed solution is based on VGG16, ResNet50 and a Vision Transformer (ViT).

In this work, we conducted rigorous experiments by fine-tuning the pretrained VGG16, ResNet50 and ViT models to classify images from the KaraAgroAI Cocoa Dataset. The images from this dataset are labelled according to three classes, namely: Healthy, Cocoa Swollen Shoot Virus Disease (CSSVD), and Anthracnose. Results indicate that our image classifiers outperform previous attempts to detect CSSVD-infected cacao plants. Below, we summarize the main contributions of this work:
\begin{enumerate}
    \item We are the first to study the CSSVD problem based on a high quality, well curated image dataset, that is, KaraAgroAI Cocoa Dataset.
    \item We fine-tuned pretrained image classification models to distinguish between cacao plants infected with CSSVD and non-infected cacao plants.
\end{enumerate}
%%%%%%%%%%%%%%%%%%%%%%%%%%%%%%
%%%%%%%%%%%%%%%%%%%%%%%%%%%%%%
\section{Related Work}
\label{sec:related work}
To describe accurately the contents within a natural scene, texture within the image plays an important role as the descriptor. Neural networks  have been very successful at learning the unique features within an image that can be used to distinguish between two images. Since the advent of convolutional neural networks, CNN for short, several image classifiers based on CNN have been proposed. Among them; CNN~\citep{10.1109/35.41400}, VGG~\citep{10.1109/35.41400}, DCNN~\citep{https://doi.org/10.48550/arxiv.1511.02136}, ResNet~\citep{https://doi.org/10.48550/arxiv.1511.02136}, AlexNet~\citep{NIPS2012_c399862d}, Inception~\citep{https://doi.org/10.48550/arxiv.1409.4842}, GoogleNet~\citep{https://doi.org/10.48550/arxiv.1409.4842}, EfficientNet~\citep{https://doi.org/10.48550/arxiv.1905.11946}, LeNet~\citep{726791} and MobileNet~\citep{https://doi.org/10.48550/arxiv.1704.04861}, have been trained to classify images which belong to plants. The main goal is to distinguish between infected and non-infected plants. 

\citep{agriculture12081192, 9417910, 9742921, app12146982} are major works attempting to detect plant diseases using contemporary image classifiers. 

Next, we describe the other works that train image classifiers to detect CSSVD in cacao plants. ~\citep{1a83bf960ebd4e4a877800854e1e591b} proposed the extraction of characteristic vectors with HoG (Histograms of Oriented Gradient), LBP (Local Binary Pattern), and the SVM (Support Vector Machine) algorithms and then trained SVM, Random Forest, and artificial neural networks to distinguish between infected and non-infected cacao plants.

To make image classification more accessible to farmers~\citep{10.1016/j.procs.2022.07.013} developed a smartphone application. Their implementation includes CNN based models trained to detect both Swollen Shoot and Black Pod diseases. Moreover,~\citep{agronomy10111642} proposed a feature extraction method for a low-computational device to classify cacao beans. Furthermore,~\citep{articleZZZ} studied the identification of infected cacao beans. However, the closest to our work is~\citep{Coulibaly} who trained a CNN for binary classification into healthy and infected cacao. Our work includes 3-class classification of cacao plants into; \emph{Anthracnose, CSSVD, and Healthy}.

In all of the above works, the use of high-quality image data remains absent. Moreover, the works mentioned above do not release their data. To the best of our knowledge, this is the first attempt to train image classifiers to detect CSSVD infected cacao plants using high-resolution and well-curated labeled image training data.
\section{Methods}
\label{sec:Methods}
In this section, we describe all the methods used to pre-process the images before they're fed to the classifiers; and the classifiers used to learn image patterns within cacao plant leaves. The main goal is to learn patterns withing the images that make it possible to determine to which of the three classes an image belongs.  Our work includes two image classifiers based on CNNs, namely: VGG16~\citep{10.1109/35.41400}, and ResNet50~\citep{https://doi.org/10.48550/arxiv.1511.02136} shown in Figure~\ref{fig:ConvNet};  and one classifier based on the vision transformer~\citep{https://doi.org/10.48550/arxiv.2010.11929}, shown in Figure~\ref{fig:ViT}.

Formally, given an input image $X$, our three-label image classifier aims to estimate the function necessary to determine the presence of labels (and categorize image $X$ into any of the labels) in the set \(\mathcal{Y}\) = \{\textit{Anthracnose, CSSVD, Healthy}\}. This can be written as follows,
$$
\begin{aligned}
f: \mathbb{R}^{w \times h} & \rightarrow  [0,1]^N \\
X \mapsto \mathbf{y} &=\left(y_i\right)_{i \in \mathcal{Y}},
\end{aligned}
$$
in which $w$ and $h$ are the pixel-wise width and height of the image, respectively. Note that $y_i=1$ if the label $i$ is present in $X$, else $y_i=0$. The set of images is \( \mathcal{X} \) = ($x_i$\ldots $x_n$); and there are 17,703 images in the KaraAgroAI Cocoa dataset.
\label{subsec:Convnets}
\subsection{VGG16} 
\label{subsec:VGG16}
We follow the same steps as reported in~\citep{https://doi.org/10.48550/arxiv.1409.1556}. \textbf{First,} the input consists of fixed-size 224 × 224 RGB images. Preprocessing of the input includes subtracting, from each pixel, the mean RGB value computed on the training set. \textbf{Next,} all images are passed through a stack of convolutional (conv.) layers, where we use filters with a 3 × 3 receptive field. The convolution stride is fixed to 1 pixel; the spatial padding of conv. layer input is such that the spatial resolution is preserved after convolution, i.e. the padding is 1 pixel for 3 × 3 conv. layers. Spatial pooling is carried out by five max-pooling layers, which follow some of the conv. layers, and not all the conv. layers are followed by max-pooling). Max-pooling is performed over a 2 × 2 pixel window, with stride 2. We adopt rectification non-linearity for each hidden layer. \textbf{Last,} a stack of convolutional layers is followed by three Fully-Connected (FC) layers: the first two have 4096 channels each. We modify the third FC layer to have 3 channels (one for each class). The final layer is the soft-max layer.
\subsection{ResNet50} 
\label{subsec:ResNet50}
For the CSSVD image classifier based on ResNet50, we adopt the configuration as described in the works of~\citep{https://doi.org/10.48550/arxiv.1512.03385}. \textbf{First,} we preprocess all input images in the same way as in the VGG16 model above, that is,  the mean RGB value computed on the training set is subtracted from each pixel. \textbf{Next,} the image goes through a stack of convolutional layers most of which have 3×3 filters and follow these two design rules: (i) for the same output feature map size, the layers have the same number of filters; and (ii) if the feature map size is halved, the number of filters is doubled so as to preserve the time complexity per layer. Then, we perform downsampling directly by convolutional layers that have a stride of 2. 

%%%%%%%%%%% Implementation for ResNet50
The implementation of ResNet50 is described as follows. The image is resized with its shorter side randomly sampled in [256, 480] for scale augmentation and a 224×224 crop is randomly sampled from an image or its horizontal flip. The standard color augmentation is used. We adopt batch normalization (BN) just after each convolution and before activation. We used Adam optimizer with a batch size of 64. The learning rate starts from 0.001 and is divided by 10 when the error plateaus, and the models are trained for up to 20 epochs. We use a weight decay of 0.0001 and a momentum of 0.9. We do not use dropout. 
\subsection{Vision Transformer (ViT)}
\label{subsec:ViT}
We adopt the vision transformer developed by~\citep{https://doi.org/10.48550/arxiv.2010.11929} for for CSSVD classification. This vision transformer model is depicted in Figure~\ref{fig:ViT}. The standard Transformer receives as input a 1D sequence of token embeddings. To handle $2 \mathrm{D}$ images, we reshape the image $\mathbf{x} \in \mathbb{R}^{H \times W \times C}$ into a sequence of flattened $2 \mathrm{D}$ patches $\mathbf{x}_p \in \mathbb{R}^{N \times\left(P^2 \cdot C\right)}$, where $(H, W)$ is the resolution of the original image, $C$ is the number of channels, $(P, P)$ is the resolution of each image patch, and $N=H W / P^2$ is the resulting number of patches, which also serves as the effective input sequence length for the Transformer. The Transformer uses constant latent vector size $D$ through all of its layers, so we flatten the patches and map to $D$ dimensions with a trainable linear projection (Eq. 1). The output of this projection is referred to as the patch embeddings.

We prepended a learnable embedding to the sequence of embedded patches $\left(\mathbf{z}_0^0=\mathbf{x}_{\text {class }}\right)$, whose state at the output of the Transformer encoder $\left(\mathbf{z}_L^0\right)$ serves as the image representation $\mathbf{y}$. Both during pre-training and fine-tuning, a classification head is attached to $\mathbf{z}_L^0$. The classification head is implemented by a MLP with one hidden layer at pre-training time and by a single linear layer at fine-tuning time.

Position embeddings are added to the patch embeddings to retain positional information. We use standard learnable 1D position embeddings. The resulting sequence of embedding vectors serves as input to the encoder.

The Transformer encoder consists of alternating layers of multi-headed self-attention (MSA) and MLP blocks. Layernorm (LN) is applied before every block, and residual connections after every block.
%%%%%%%%%%%%%%%%%%%%%%%%%%%%
\begin{table*}[t!]
\centering
\scriptsize
\caption{Summary of per-class classifier performance (\%) on KaraAgroAI Dataset. The number of epochs is 20. For all experiments, the Image\_size is 224X224. In addition, we report the overall precision, recall and F1-score resp. for all classes. }
\begin{tabular}{|l|l|l|l|l|l|l|l|l|l|l|l|l|}
    \hline
    Model   & \multicolumn{4}{|c|}{}        & \multicolumn{4}{|c|}{}     & \multicolumn{4}{|c|}{}\\
            & \multicolumn{4}{|c|}{ \textbf{VGG16}}  & \multicolumn{4}{|c|}{ \textbf{ResNet50}} & \multicolumn{4}{|c|}{ \textbf{Vision Transformer}}\\
    \hline
    \multicolumn{1}{|c|}{Class} & Precision & Recall & F1    & Acc & Precision & Recall & F1    & Acc & Precision & Recall & F1    & Acc \\
    \hline
    Anthracnose               & 98.21     & 85.08  & 91.17 & --  & 98.66     & 85.66  & 91.70 & --  & 90.76     & 62.79  & 74.22 & --  \\
    CSSVD                     & 87.39     & 99.17  & 92.91 & --  & 90.64    & 100  & 95.09 & --  & 74.0      & 99.31  & 84.81 & --  \\
    Healthy                   & 96.54     & 90.996 & 93.69 & --  & 96.89    & 95.59  & 96.24 & --   & 86.05     & 71.01  & 77.81 & -- \\
    \hline
    Overall                   & 96.54     & 90.996 & 93.69 & 92  &95.39     & 93.75  & 94.34 & 94  & 86.05     & 71.01  & 77.81 & 80 \\
 \hline
 \end{tabular}
 \label{Table: Experiments_per_class_performance}
\end{table*}
%%%%%%%%%%%%%%%%%%%%%%%%%%%%
\section{Experiments}
\label{sec:Experiments}
In this section, we report the results obtained on KaraAgroAI Cocoa Dataset~\cite{DVN/BBGQSP_2022}. \\
% \vspace{0.1}
\subsection{KaraAgroAI Dataset}
\label{subsec:KaraAgroAI Dataset}
KaraAgroAI Cocoa Dataset is a multi-label image dataset that provides images across three classes, namely; Healthy, Cocoa Swollen Shoot Virus Disease (CSSVD), and Anthracnose. We gathered all the data and the per-class statistics are shown in the Table~\ref{Table: Experiments_KaraAgroAI}.
\begin{table}[h!]
\centering
\footnotesize
\caption{Number of images per class in KaraAgroAI Dataset.}
\begin{tabular}{ |c|c|c|c|c| }
 \hline
       & {\bf Anthracnose} & {\bf CSSVD} & {\bf Healthy} & {\bf Total}\\
 \hline
 {\bf \#Images} &  5,162 &  7,292 &  5,249 & 17,703\\
 \hline
 \end{tabular}
 \label{Table: Experiments_KaraAgroAI}
\end{table}
%%%%%%%%%%%%%%%%%%%%%%%%%%%%
To train the image classifiers, we split the dataset into train, validation and test sets (we used the 80/10/10 split resp.). The number of samples per class; in the train, validation and test sets; is shown in Table~\ref{Table: Experiments_Train_Val_Test}.
\begin{table}[h!]
\centering
\footnotesize
\caption{KaraAgroAI data split for experiments.}
\begin{tabular}{ |c|c|c|c|c| }
 \hline
       & {\bf Train} & {\bf Validation} & {\bf Test} & {\bf Total}\\
 \hline
 Anthracnose    & 4,131 & 515   &  516  & 5,162\\ 
 CSSVD          & 5,838 & 727   &  727  & 7,292 \\ 
 Healthy        & 4,207 & 521   &  521  & 5,249\\ 
 \hline
 {\bf \#Images} & {\bf 14,176} & {\bf 1,763} & {\bf 1,764} & 17,703\\
 \hline
 \end{tabular}
 \label{Table: Experiments_Train_Val_Test}
\end{table}
%%%%%%%%%%%%%%%%%%%%%%%%%%%%
\subsection{Experiment Settings} 
\label{subsec:Experiment Settings}
We selected three image classification methods, namely: VGG16, ResNet50 and ViT.

\textbf{Hyper-parameter tuning:} Our proposed classifiers are optimized with Adam optimizer, and the optimizer learning rate starts from 1e-3, decaying two times smaller every epoch. The total number of epochs is 20 with proper early stopping. The batch size is 64. The input of each dataset is zero-mean normalized. All hyper-parameters are shown in Table~\ref{Table: Experiments_Hyperparameter}.

\textbf{Computation server:} All the models were trained and tested on a single Intel® Xeon® Platinum 8160 96-core CPU with 1.53TB RAM; running Ubuntu 20.04.5.\\
\begin{table}[h!]
\centering
\footnotesize
\caption{Model Hyperparameters. }
\begin{tabular}{ |c|c|c|c| }
 \hline
   {\bf Hyperparameter}    & {\bf VGG16}  & {\bf ResNet50} & {\bf Vision Transformer} \\
 \hline
 Input image\_size              & 224x224    & 224x224   & 224x224\\
 Adam $\beta_1$                 & 0.0        &  0.9      & 0.9\\ 
 Adam $\beta_2$                 & 0.999      & 0.999     & 0.999 \\ 
 Batch Size                     & 64         & 64        & 64\\
 Dropout, Attention             & N/A        & N/A       & 0.5\\
 Dropout, Feedforward           & 0.5        & N/A       & 0.2\\
 Encoder Layer Used             & CNN        & CNN       & CNN\\
 % Gradient Accumulation Steps    & 0.         & 0         & 0 \\
 Hid Dim Class Head             & N/A        & N/A       & 256 \\
 Number of heads                & N/A        & N/A       & 8\\
 Embed\_size                    & 256        & 256       & 256 \\
 LR Scheduler                   & Linear    & Linear     & Linear\\
 % Step\_size                     & 7          & 7         & 7 \\
 Learning Rate                  & 1e-2       & 1e-3      & 1e-3\\
 Momentum                       & 0.9        & 0.9       & 0.9 \\
 Weight Decay                   & 0.0005     & 0.0001    & 0.03 \\
 Total Epochs                   & 20         & 20        & 20\\
 \hline
 \end{tabular}
 \label{Table: Experiments_Hyperparameter}
\end{table}
%%%%%%%%%%%%%%%%%%%%%%%%%%%%
\subsection{Results and Analysis}
\label{subsec:Results and Analysis}
We report the precision, recall, and F1-score per class; for all the image classifiers. Moreover, we report the number of model parameters on KaraAgroAI Cocoa dataset, shown in Table~\ref{Table: Experiments_model_size_mAP}. 
\begin{table}[h!]
\centering
\footnotesize
\caption{Summary of models used, parameter size and the size of each model on KaraAgroAI dataset. }
\begin{tabular}{ |c|c|c| }
 \hline
   {\bf Method}    & {\bf \#Parameters} & {\bf Model size (MBs)} \\
 \hline
 VGG16    & 138.3M  &  14,469.3\\ 
 ResNet50 & 23.5M  & 11,513.3 \\ 
 ViT      & 6.8M & 4,797.2 \\
 \hline
 \end{tabular}
 \label{Table: Experiments_model_size_mAP}
\end{table}
%%%%%%%%%%%%%%%%%%%%%%%%%%%%
\begin{table}[h!]
\centering
\footnotesize
\caption{Summary of Classification Accuracy on KaraAgroAI dataset. }
\begin{tabular}{ |c|c|c|c| }
 \hline
   {\bf Model} & {\bf VGG16} & {\bf ResNet50}  & {\bf ViT}\\
 \hline
 Overall Accuracy(\%) & 92  &  94 & 80\\ 
%  CSSVD        & 99M  & 98.8 \\ 
%  Healthy      & 200M & 87.7 \\
 \hline
 \end{tabular}
 \label{Table: Experiments_Overall_Accuracy}
\end{table}
%%%%%%%%%%%%%%%%%%%%%%%%%%%%
We compare the best-performing model on CSSVD image classification with the recent work of~\citep{Coulibaly}. A comparison of their best-performing model (as reported in their paper) and ours is shown in Table~\ref{Table: Experiments_BESTmodel_vs_Related_work}.
\begin{table}[h!]
\centering
\scriptsize
\caption{Comparison for overall scores between our work, and related work on CSSVD classification. }
\begin{tabular}{ |c|c|c|c|c|c|c|c| }
 \hline
   {\bf Method}    & {\bf Base Model} & {\bf Precision} & {\bf Recall} & {\bf F1} & {\bf Accuracy}\\
 \hline
 ~\cite{Coulibaly}  & CNN         & 74    & 84     & 98        & 88\\ 
 \hline
 \hline
 \textbf{Ours}      & ViT         & 86.05    & 71.01    & 77.81    & 80\\
 \hline
 \textbf{Ours}      & VGG16       & 93.86    & 91.41   & 92.67    & 92\\
 \hline
 \textbf{Ours (best)}      & ResNet50   & {\bf 95.39}   & {\bf 93.75}   & {\bf 94.34}     & {\bf 94}\\
 \hline
 \end{tabular}
 \label{Table: Experiments_BESTmodel_vs_Related_work}
\end{table}
\section{Conclusion}
\label{Conclusion}
In this work, we investigated the ability for pre-trained image classification models to detect the rampant \emph{cacao swollen shoot virus disease} (that is, \emph{CSSVD}) in cacao plants. We leveraged the image-pattern recognition ability of \emph{ResNet50, VGG16}, and \emph{Vision Transformer} to identify infected cacao plants when trained with images of cacao leaves, stems and fruits. 

We evaluated the classifiers on a recently released and publicly accessible KaraAgroAI Cocoa dataset comprising 17,703 images. 

Experiment results indicate that ResNet50 and VGG16 are sufficient to classify cacao images among the three classes, that is, \emph{Anthracnose, CSSVD} and \emph{Healthy}; when trained for only 20 epochs. 

Our best image classifier (that is, ResNet50) achieves 95.39\% precision, 93.75\% recall, 94.34\%F1 and 94\% accuracy on only 20 epochs. There is a +9.75\% improvement in recall when compared to previous works. 

Our results indicate that the image-classifiers learn to identify cacao plants infected with CSSVD when trained with images of cacao leaves, stems and fruits. 

\vfill\pagebreak

\bibliographystyle{IEEEbib}
\bibliography{main}

\end{document}